\documentclass[runningheads,anonymous]{llncs}
\usepackage{graphicx}
\usepackage{hyperref}
\usepackage{amsmath}
\usepackage{booktabs}
\usepackage{breakcites}
\usepackage[nodisplayskipstretch]{setspace}
\usepackage[table,xcdraw]{xcolor}
\usepackage[normalem]{ulem}
\useunder{\uline}{\ul}{}
\begin{document}

\title{LongStory: Coherent, Complete and Length Controlled Long story Generation }
\titlerunning{LongStory}

\author{Kyeongman Park \and
Nakyeong Yang \and
Kyomin Jung\thanks{Corresponding Author.}}
\authorrunning{Park et al.}
\vspace{-5px}
\institute{Seoul National University, Seoul, Republic of Korea \\ \email{\{zzangmane,yny0506,kjung\}@snu.ac.kr}}

\newcommand{\yny}[1]{\textcolor{orange}{[--yny: #1]}}

\maketitle              % typeset the header of the contribution
\vspace{-25px}
\begin{abstract}
A human author can write any length of story without losing coherence. Also, they always bring the story to a proper ending, an ability that current language models lack. In this work, we present the LongStory for coherent, complete, and length-controlled long story generation.
LongStory introduces two novel methodologies: (1) the long and short-term contexts weight calibrator (CWC) and (2) long story structural positions (LSP). The CWC adjusts weights for long-term context Memory and short-term context Cheating, acknowledging their distinct roles. The LSP employs discourse tokens to convey the structural positions of a long story.
Trained on three datasets with varied average story lengths, LongStory outperforms other baselines, including the strong story generator Plotmachine, in coherence, completeness, relevance, and repetitiveness. We also perform zero-shot tests on each dataset to assess the model's ability to predict outcomes beyond its training data and validate our methodology by comparing its performance with variants of our model.
\vspace{-5px}
\end{abstract}
\vspace{-25px}
\section{Introduction}
\vspace{-10px}
The story generation task is one of the most challenging problems in natural language processing since it requires writing long lengths with a consistent context.
Existing studies have tried to solve this problem but have failed to generate variable lengths of stories since they have only considered fixed lengths when generating stories.
 
Longformer\cite{Beltagy_2020} has tried to solve this problem of handling longer sequences by combining global and local attention in sliding windows. However, they only focused on increasing the input context window size. Their maximum generation length is only 1,024 tokens. Short-length problems in text generation also occur in recent large language models such as GPT-4\cite{OpenAI_2023}. Plug-and-blend\cite{Lin_2021} has also covered a similar problem, but the length limitation remains since it only aimed to control a few sentences.

To address this challenge, a recursive paragraph generation approach is necessary to compose long stories, given the length limitations imposed by existing language models. However, this recursive generation of stories may cause undesirable forgetting of the previous context since the information leak may occur in the recursive process of information transfer (\textit{coherence})\cite{Yang_2022_DOC}.
Also, existing studies have mentioned that language models tend to repeat the same story in a recursive generation setting. (\textit{variety} and \textit{repetitiveness})\cite{McCoy_2023,Rashkin_2020}.

Thus, previous works \cite{Wang_2020,Yang_2022_DOC,Rashkin_2020} have tried to define coherence and repetitiveness metrics and used them to measure the ability of story generators.

However, their attempts are also limited only to short sentence generation problems and have never considered \textbf{completeness}. Completeness is the ability to conclude a story of any length properly. It is a significant metric not only for evaluating story generation but also for a wide range of open-domain generation and dialogue system tasks. Therefore, in our model evaluation, we consider not only coherence and repetitiveness but also prioritize completeness as a critical measure of performance.

In this paper, we tackle this challenging problem by introducing a novel long story generator, \textit{Coherent, Complete and Length Controlled Long story Generator}(LongStory), which covers from a few hundred tokens to 10k tokens length, not limited to only a few sentences. The LongStory utilizes two novel methods: (1) long and short-term contexts weight calibrator (CWC) and (2) long story structural positions (LSP) to tackle coherence and completeness.

Specifically, we implement the CWC using BERT-tiny to calibrate the weight to which long-term and short-term contexts are utilized since both contexts contribute differently in every paragraph when writing a story. For the LSP, we use discourse tokens, which means the order of paragraphs, to give information about the structural positions of paragraphs to the story generator (e.g., $<front>$, $<middle>$,$<ending>$, $<next\_is\_ending>$).
Our model uses more abundant discourse tokens than the previous study \cite{Rashkin_2020} since a detailed understanding of story structure is essential to generate much longer posts.

We use three diverse story generation datasets with varying average lengths to train our model on representations of different story lengths. We introduce quantitative metrics for coherence, completeness, and repetitiveness, evaluating the model's performance against other baselines, including the established story generator Plotmachine\cite{Rashkin_2020}.
The experimental results for three story generation datasets demonstrate that our model outperforms the other baselines in coherence, completeness, and relevance. Surprisingly, our model also shows better results in repetitiveness, suggesting that our methods are effective for the variety. Furthermore, we performed zero-shot tests on each dataset to assess how well our model predicts outcomes in settings beyond its training data. Additionally, our analysis of the augmented CWC version suggests that elevating relevance does not always translate to improvements in coherence and completeness.

The main contributions of this paper are (1). a new open-domain metric called completeness. (2) a new challenge, \textit{Coherent, Complete and Length Controlled Long story Generation} (LongStory), incorporating CWC and LSP methodologies, and (3) the presentation of datasets with varying average document lengths along with zero-shot tests for them.
\vspace{-10px}
\section{Related Works}
\vspace{-10px}
Many contemporary automatic story generation models have employed prompt engineering strategies, manipulating input prompts for large pretrained language models like GPT-3\cite{Yang_2022_Re3,Lin_2021,Yang_2022_DOC}. Despite the effectiveness of these black-box models, they could not directly enhance the internal structure for optimized performance. In contrast, our approach focuses on directly improving the structural aspects of existing language models. 
\vspace{-10px}
\subsection{Neural Story Generation}
Story generators using neural networks have been developed in various ways \cite{Fan_2018,Brown_2020,Hu_2022,Guan_2020,Zellers_2019}. The main approaches have included outline-based generation \cite{Fan_2018,Rashkin_2020,Yang_2022_Re3,Yang_2022_DOC,Hu_2022,Sakaguchi_2021}, event graph-based \cite{Tang_2022_Etrica,Yao_2019}, goal-oriented methodologies \cite{Alabdulkarim_2021_goal,Pradyumna_2019}, and common sense reasoning \cite{Guan_2020,Peng_2021,Lewis_2020}. Outline-based generation methods often involve interpolating plot keywords\cite{Rashkin_2020,Hu_2022}. Our model also uses keywords to plan plots and create relevant stories. Event graph-based approaches have proven highly effective, but only if given a detailed plot of the entire article. Our model uses keywords for the entire story but not the plot of the story, thus performing a more challenging task. Goal-oriented methodologies have aimed to make the story's characters achieve given goals, akin to controlling the ending of a story\cite{Peng_2018,Wang_2020}. Common sense inference methods have focused on generating realistic texts by training on common sense datasets. 
\vspace{-10px}
\paragraph{Length-Controllable Story Generation.}
 Many models have generated text within a fixed range of length. The length has been mostly determined when selecting the training datasets. For example, Plotmachine \cite{Rashkin_2020} has fixed the number of paragraphs it generates, while EtriCA \cite{Tang_2022_Etrica} and Fusion \cite{Fan_2018} also have set their generation lengths by their training dataset. Re3 \cite{Yang_2022_Re3} has generated narratives by applying pre-trained GPT-3 \cite{Brown_2020}, and the length has been adjusted by manipulating the prompts to GPT-3. However, Re3 has not solely focused on controlling length. As such, it has not been specifically designed to ensure that the models generate without losing coherence and completeness with various lengths. 
\vspace{-10px}
 \subsection{Recursive Models}
Due to finite parameters, there has always been a limit to the length a language model can generate at once. While Longformer\cite{Beltagy_2020} and GPT-4\cite{OpenAI_2023} have introduced 5K or even 100K context windows, their output length remains only a few thousand tokens at most. In contrast, human writers can produce thousands of pages without encountering such limitations. This reality implies the necessity of employing a language model recursively \cite{Rashkin_2020,Yang_2022_Re3,Hu_2022}. 
\vspace{-10px}
\subsection{Autometic metrics}
In NLP generation tasks, traditional evaluation methods commonly involve metrics such as ROUGE \cite{Lin_2004}, BLEU \cite{Papineni_2002}. These metrics predominantly rely on n-gram matching between labels and predictions. However, applying these evaluation methods to the story generation task may not be sufficient. The reason lies in the fact that a well-crafted story may not necessarily exhibit similarity to the label; in fact, a story resembling the label might score poorly in terms of diversity and creativity \cite{Safovich_2020,Yang_2021}. In our case, we additionally employ \textit{coherence}, \textit{completeness}, and \textit{repetitiveness} as the primary evaluation metrics along with the ROUGE scores. While \textit{coherence} and \textit{repetitiveness} is a well-established metric in many natural language generation tasks\cite{Peng_2018,Rashkin_2020,Yang_2022_Re3,Yang_2022_DOC,Li_2019,Hu_2022}, we are the first to use \textit{completeness} as a metric. 
\vspace{-10px}
\setcounter{secnumdepth}{3}
\section{Methodology}
\begin{figure}
\includegraphics[width=0.9\textwidth]{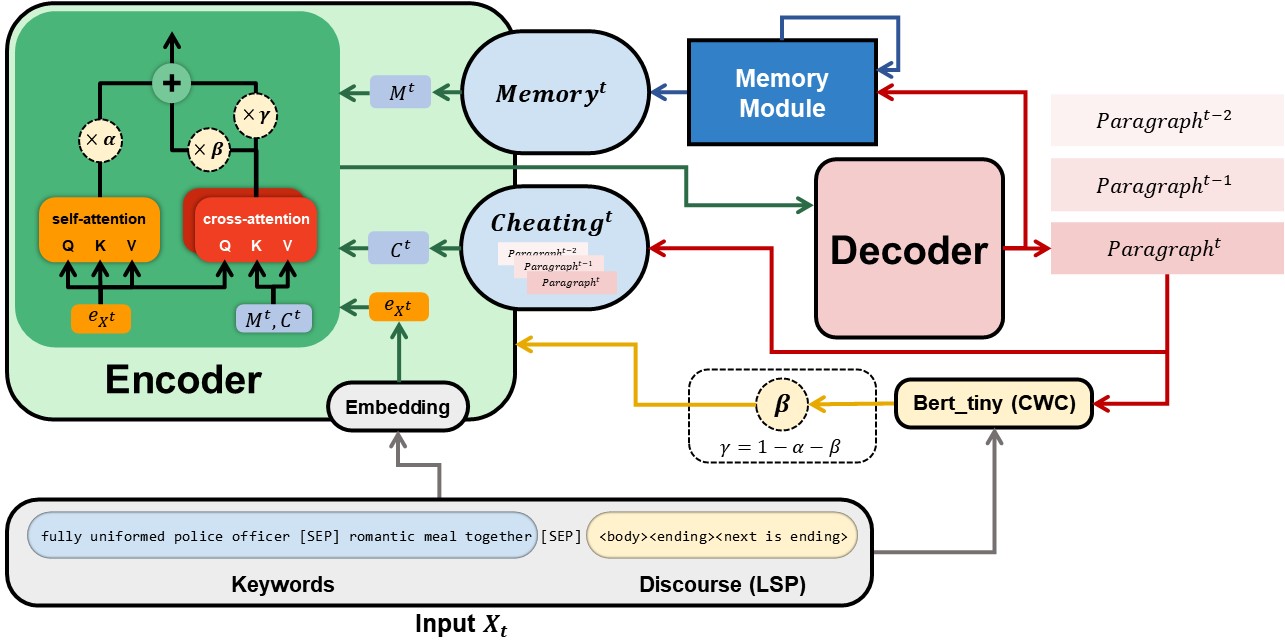}
\centering
\caption{Model architecture. LongStory takes the keywords of the entire story and the discourse tokens(LSP) representing the order of the target paragraphs as input. The BERT-tiny serves as the long and short-term context weight calibrator (CWC), determining the degree to which long-term and short-term contexts are employed. The CWC takes the discourse tokens and the last generated paragraph as inputs and outputs the optimal $\beta$ and $\gamma$(defined as 1-$\alpha$-$\beta$) for every paragraph. While the $\alpha$ is a hyperparameter applied to input embedding, $\beta$ is a learnable parameter for the long-term context \textit{Memory}($M^t$) and short-term context \textit{Cheating}($C^t$)} \label{fig1}
\end{figure}
\vspace{-30px}
\subsection{Task Description}
Our challenge can be described as follows : 
Given keywords K=($k_1$,$k_2$,$k_3$...), discourse tokens $D_t$=($d_1^t$,$d_2^t$,$d_3^t$...) (each $k_i$,$d_i^t$ is a token), and a separation token [SEP], an input $X^t$=($k_1$,[SEP],$k_2$,[SEP],$k_3$,...,[SEP],$d_1^t$,$d_2^t$,$d_3^t$...). The model should generate each t-th paragraph recursively: $Y^t$=($y_1^t$,$y_2^t$,$y_3^t$,...) (each $y_i^t$ is a sentence, a set of tokens) from given contexts of previous paragraphs and the input $X^t$, which is coherent and completive. The model should minimize negative log-likelihood loss $L_{LM}$ of each paragraph t :
\vspace{-10px}
\begin{equation}
L_{LM}=-\sum_{i=1}^{n} log P(y_j^t|{y^t_{<j}} , X^t)\
\end{equation}
\vspace{-10px}
\begin{equation}
log P(y_j^t|{y^t_{<j}} , X^t) = softmax(H_j^tW+b)
\end{equation}
\vspace{-10px}
\begin{equation}
H_j^t = LM(X^t, M^t, C^t)\label{eq1}
\end{equation}

where $LM$ is an attention-based language model, $H_j^t$ is the j-th position of the language model's last hidden state, $M^t$ is the \textit{Memory} which saves the long-term context by the t-th paragraph, and $C^t$ is \textit{Cheating} which keeps the short term context from the very last paragraphs it generated. The $M^t$ ensures a sustained understanding of the broader context, while $C^t$ contributes to short-term continuity.  
\vspace{-10px}
\subsection{Long and short term contexts weight calibrator(CWC)}
BERT-tiny is inserted into our model as a calibrator and trained on the same negative log likelihood loss, $L_{LM}$. The pooler output of BERT-tiny is used for the determination of the context weight, $\beta,\gamma$.
\vspace{-5px}
\begin{equation}
B_k^t=Bert(D^t,Y^{t-1})
\end{equation}
\vspace{-10px}
\begin{equation}
\quad\beta=(1-\alpha)\cdot\sigma(B_{1}^tW+b),\quad\gamma=1-\alpha-\beta
\end{equation}

where $\alpha\in[0,1]$ is a hyperparameter\footnote{We covered a test where alpha is also a learnable parameter rather than a constant hyperparameter in sec.\ref{4.3.2}. In this test, the BERT-tiny determines $\alpha$, $\beta$, and $\gamma$ independently, and finally divides by their sum, so that each variable adds up to 1. }, $\beta$ is a learnable parameter for the long-term context \textit{Memory}($M^t$) and short-term context \textit{Cheating}($C^t$), $\sigma$ is the sigmoid function, $B_k^t$ is the k-th position of the last hidden state of BERT-tiny, and $B_1^t$ is the position of $[CLS]$ token of it, often called the `pooler output.' The long and short-term contexts, $M^t$ and $C^t$, are computed as :
% \vspace{-5px}
\begin{equation}
g^t=\sigma(W_1M^{t-1}+W_2h^{t-1}),\quad \hat{M^t}=tanh(W_3M^{t-1}+W_4h^{t-1})
\end{equation}
\vspace{-15px}
\begin{equation}
M^t=(1-g^t)\odot M^{t-1} + g^t\odot \hat{M^t}
\end{equation}
\label{eq_3}
\vspace{-10px}
\begin{equation}
C^t=tanh(Emb(y_{n-c}^{t-1},...,y_{n-1}^{t-1},y_{n}^{t-1}))
\end{equation}

where $Emb$ denotes a positional embedding layer that involves the element-wise sum of word embeddings and positional encoding vectors, $h^{t-1}$ is the average embedding of $Y^{t-1}$, and $c$ is a hyperparameter determining the size of the cheating window. We utilized the same computation process of initializing and updating $M^t$ as introduced in Plotmachine. Finally, we define the computation of an attention block within a $LM$ as follows :
% \vspace{-5px}
\begin{equation}
\begin{split}
LM(X^t,M^t,C^t)=\alpha Attention(Q,K,V)&+\beta Attention(Q,M^t,M^t) \\&+\gamma Attention(Q,C^t,C^t)
\end{split}
\label{eq_2}
\end{equation}

 where Q=$Emb$($X^t$)$W^Q$, K=$Emb$($X^t$)$W^K$, V=$Emb$($X^t$)$W^V$, and \textit{Attention} is a typical transformer attention block\cite{Vaswani_2017}. By weighting the two contexts, the model can generate cohesive and contextually connected text.
\vspace{-10px}
\subsection{Long story structural positions (LSP)}
\vspace{-3px}
 We must provide structural positional information for each paragraph\cite{Guan_2021}. All discourse tokens are of the following types $<intro>$, $<body>$, $<tail>$, $<front>$, $<middle>$, $<ending>$, $<next\_is\_ending>$. $<intro>$ corresponds to the first paragraph, $<tail>$ to the last paragraph, and $<body>$ to everything in between as introduced in the previous work\cite{Rashkin_2020}. However, since our task is to generate much more paragraphs, $<body>$ token is not sufficient to give representation to our model. Therefore, four tokens ($<front>$, $<middle>$, $<ending>$, $<next-is-ending>$) are added : $<front>$ for the first 1/3, $<middle>$ for the next 1/3, and $<ending>$ for the last 1/3. $<next-is-ending>$ is for the paragraph immediately before the end, as its name implies.
\vspace{-15px}
\subsection{Base Pretrained Model}
\vspace{-5px}
We finetune a Pretrained Language Model(PLM) based on Transformer\cite{Vaswani_2017} for this task. Although our algorithm is model-agnostic so that it can apply any of PLM, we choose BART\cite{Lewis_2019} because (1) it had learned denoising sequences to efficiently extract information from a given input, appropriate for our task as the model should extract representations of the inputs and contexts. (2) it is free to download from online\footnote{\url{https://huggingface.co/facebook/bart-large}}. However, our model holds the potential for even more excellent performance when applied to larger language models in the future.
\vspace{-10px}
\section{Experiments}
\vspace{-10px}
\subsection{Experiments Set-up}
\vspace{-5px}
\subsubsection{Datasets.}
We train our models on Writing Prompts\cite{Fan_2018}, Booksum\cite{Kryściński_2021}, and Reedsy Prompts\footnote{For more information, \url{https://huggingface.co/datasets/Iyan/reedsyPrompts.}} datasets. For Booksum, we use the `Chapter' column as a dataset. Since our task is to generate various lengths of stories with coherence and completeness, we must prepare datasets of different lengths. We use the same train/valid/test split from the original papers\cite{Fan_2018,Kryściński_2021}, but since Reedsy Prompts is not used in any other works, we split it into train/valid/test sets of 60k/4k/4k.
The NLTK library's sentence tokenizer is utilized for document segmentation into sentences during truncation, with each sentence sequentially compiled into a paragraph, ensuring a maximum length of 200 tokens per paragraph.
For each dataset, we extract keywords using the RAKE algorithm\cite{Rose_2010} to make keyword-story pairs. These keywords are used as input for the model to write each paragraph of a story.
\begin{table}[]
\centering
\resizebox{0.9\textwidth}{!}{% 
\begin{tabular}{|l|r|r|r|}
\hline
                        & \multicolumn{1}{l|}{\textbf{Avg \# of Tokens}} & \multicolumn{1}{l|}{\textbf{Avg \# of Paragraphs}} & \multicolumn{1}{l|}{\textbf{\# of documents}} \\ \hline
\textbf{WritingPrompts} & 768                                            & 3.4                                                & 300k                                              \\ \hline
\textbf{Booksum}        & 6065                                           & 27.7                                               & 12k                                               \\ \hline
\textbf{ReedsyPrompts}  & 2426                                           & 12.3                                               & 68k                                               \\ \hline
\end{tabular}%
}
\vspace{5px}
\caption{The average number of tokens, paragraphs per document, and the number of documents used in experiments.}\label{tab1}
\end{table}
\vspace{-25px}
\subsubsection{Baselines.}
We mainly compare our model to Plotmachine, the only model that performs paragraph-level recursive story generation. We also compare the performance of our model against GPT-2\cite{Budzianowski_2019} and BART with the recursive generation setting, no applied memory, and cheating contexts; thus, it does not utilize the CWC. Plotmachine receives only the discourse tokens from the original paper, while other baselines are given the same as ours.
We also compare our model with the ablated version. The \textit{LongStory with no memory} (LongStory$^{\neg M}$), \textit{LongStory with no cheating} (LongStory$^{\neg C}$) and \textit{LongStory with no new discourse} (LongStory$^{\neg D}$) show the effectiveness of context vectors and newly added discourse tokens.
\vspace{-20px}
\subsubsection{Implementation Details.}
Experiments are run on an NVIDIA RTX A5000 GPU server. For training steps, we use the teacher-forcing method so each label is used for not only $Y_t$ but also contexts. We set the hyperparameter $\alpha=\frac{1}{2}$ because providing the model with sufficient inductive bias for input embeddings was beneficial for coherence and completeness. We also configured the size of the cheating window $c$ up to the last three paragraphs it generated. For generation, we use beam sampling with \textit{top\_k}=50, \textit{top\_p}=0.95, \textit{no\_repeat\_ngram\_size}=3 and \textit{repetition\_penalty}=3.5\cite{Guan_2020} for every length of paragraphs. To reproduce Plotmachine, we refer to the author's github repository.\footnote{\href{https://github.com/hrashkin/plotmachines}{\url{https://github.com/hrashkin/plotmachines}}}.  
% \vspace{-20px}
\begin{table}
\centering
\resizebox{\textwidth}{!}{
\begin{tabular}{lcclccccccc}
\multicolumn{11}{c}{\cellcolor[HTML]{EFEFEF}{\color[HTML]{3166FF} \textbf{5 Paragraphs}}}                                                                                                                                                                                       \\ \hline
\multicolumn{1}{l|}{}                           & \textbf{R-1}             & \textbf{R-2}             & \textbf{R-L}             & \textbf{B2}   & \textbf{B-3}  & \textbf{B-4}  & \textbf{B-5}  & \textit{\textbf{Coherence}} & \textit{\textbf{Completeness}} & \textbf{AvgL} \\ \cline{1-3} \cline{5-11} 
\multicolumn{1}{l|}{\textbf{Golden Label}}      & \cellcolor[HTML]{656565} & \cellcolor[HTML]{656565} & \cellcolor[HTML]{656565} & 0.28          & 0.15          & 0.07          & 0.03          & 82.62                       & 62.5                           & 161           \\ \cline{1-3} \cline{5-11} 
\multicolumn{1}{l|}{\textbf{Plotmachine}}       & 29.85                    & 7.15                     & 28.16                    & \textbf{0.47} & \textbf{0.33} & \textbf{0.23} & \textbf{0.16} & 60.19                       & 2.47                           & 192           \\
\multicolumn{1}{l|}{\textbf{BART}}              & 38.77                    & 13.64                    & 37.35                    & 0.61          & 0.49          & 0.41          & 0.33          & 54.93                       & 2.39                           & 167           \\
\multicolumn{1}{l|}{\textbf{GPT-2}}             & 33.41                    & 9.25                     & 31.82                    & 0.55          & 0.42          & 0.32          & 0.24          & {\ul 63.08}                 & 1.13                           & 193           \\ \hline
\multicolumn{1}{l|}{\textbf{LongStory}}              & 39.41                    & 13.50                    & 37.84                    & {\ul 0.49}    & {\ul 0.36}    & {\ul 0.27}    & {\ul 0.20}    & \textbf{65.13}              & \textbf{21.15}                 & 158           \\
\multicolumn{1}{l|}{\textbf{$LongStory^{\neg M}$}}       & {\ul 39.69}              & 13.66                    & \textbf{38.20}           & 0.51          & 0.38          & 0.29          & 0.22          & 59.38                       & -3.04                          & 167           \\
\multicolumn{1}{l|}{\textbf{$LongStory^{\neg C}$}}     & 39.35                    & \textbf{14.02}           & 37.90                    & 0.62          & 0.51          & 0.42          & 0.35          & 62.59                       & 8.53                           & 170           \\
\multicolumn{1}{l|}{\textbf{$LongStory^{\neg D}$}} & \textbf{39.69}           & {\ul 13.67}              & {\ul 38.08}              & 0.50          & 0.37          & 0.28          & 0.21          & 62.27                       & {\ul 9.21}                     & 166           \\ \hline
\multicolumn{11}{c}{\cellcolor[HTML]{EFEFEF}{\color[HTML]{3166FF} \textbf{10 Paragraphs}}}                                                                                                                                                                                      \\ \hline
\multicolumn{1}{l|}{}                           & \textbf{R-1}             & \textbf{R-2}             & \textbf{R-L}             & \textbf{B2}   & \textbf{B-3}  & \textbf{B-4}  & \textbf{B-5}  & \textit{\textbf{Coherence}} & \textit{\textbf{Completeness}} & \textbf{AvgL} \\ \hline
\multicolumn{1}{l|}{\textbf{Golden Label}}      & \cellcolor[HTML]{656565} & \cellcolor[HTML]{656565} & \cellcolor[HTML]{656565} & 0.37          & 0.19          & 0.08          & 0.03          & 80.44                       & 55.5                           & 164           \\ \hline
\multicolumn{1}{l|}{\textbf{Plotmachine}}       & 30.34                    & 7.74                     & 28.87                    & 0.62          & 0.47          & {\ul 0.34}    & {\ul 0.25}    & 64.17                       & 3                              & 195           \\
\multicolumn{1}{l|}{\textbf{BART}}              & 36.28                    & 11.71                    & 34.83                    & 0.67          & 0.53          & 0.43          & 0.34          & 60.46                       & 2.65                           & 172           \\
\multicolumn{1}{l|}{\textbf{GPT-2}}             & 32.32                    & 8.84                     & 30.74                    & 0.64          & 0.49          & 0.36          & 0.27          & \textbf{66.06}              & 3.85                           & 196           \\ \hline
\multicolumn{1}{l|}{\textbf{LongStory}}              & {\ul 37.46}              & {\ul 12.11}              & {\ul 35.82}              & \textbf{0.59} & \textbf{0.43} & \textbf{0.31} & \textbf{0.23} & {\ul 64.8}                  & \textbf{21.15}                 & 166           \\
\multicolumn{1}{l|}{\textbf{$LongStory^{\neg M}$}}       & 37.32                    & 12.05                    & 35.72                    & {\ul 0.61}    & {\ul 0.45}    & 0.34          & 0.25          & 58.4                        & 3.59                           & 171           \\
\multicolumn{1}{l|}{\textbf{$LongStory^{\neg C}$}}     & 37.06                    & 12.11                    & 35.49                    & 0.67          & 0.53          & 0.43          & 0.34          & 63.24                       & 6.56                           & 170           \\
\multicolumn{1}{l|}{\textbf{$LongStory^{\neg D}$}} & \textbf{37.54}           & \textbf{12.30}           & \textbf{35.86}           & 0.60          & 0.44          & 0.33          & 0.25          & 63.4                        & {\ul 11.12}                    & 171           \\ \hline
\multicolumn{11}{c}{\cellcolor[HTML]{EFEFEF}{\color[HTML]{3166FF} \textbf{20 Paragraghs}}}                                                                                                                                                                                      \\ \hline
\multicolumn{1}{l|}{}                           & \textbf{R-1}             & \textbf{R-2}             & \textbf{R-L}             & \textbf{B2}   & \textbf{B-3}  & \textbf{B-4}  & \textbf{B-5}  & \textit{\textbf{Coherence}} & \textit{\textbf{Completeness}} & \textbf{AvgL} \\ \hline
\multicolumn{1}{l|}{\textbf{Golden Label}}      & \cellcolor[HTML]{656565} & \cellcolor[HTML]{656565} & \cellcolor[HTML]{656565} & 0.45          & 0.24          & 0.12          & 0.05          & 80.13                       & 57.66                          & 166           \\ \hline
\multicolumn{1}{l|}{\textbf{Plotmachine}}       & 31.45                    & 9.49                     & 29.96                    & 0.71          & 0.57          & 0.44          & 0.34          & 64.26                       & 4.75                           & 194           \\
\multicolumn{1}{l|}{\textbf{BART}}              & 35.43                    & 12.29                    & 34.02                    & 0.74          & 0.62          & 0.51          & 0.42          & 61.37                       & -1.53                          & 169           \\
\multicolumn{1}{l|}{\textbf{GPT-2}}             & 33.28                    & 10.52                    & 31.60                    & 0.72          & 0.58          & 0.46          & 0.35          & \textbf{65.8}               & 0.7                            & 198           \\ \hline
\multicolumn{1}{l|}{\textbf{LongStory}}              & {\ul 36.85}              & 12.69                    & {\ul 35.22}              & \textbf{0.68} & \textbf{0.53} & \textbf{0.40} & \textbf{0.31} & {\ul 64.6}                  & \textbf{6.73}                  & 169           \\
\multicolumn{1}{l|}{\textbf{$LongStory^{\neg M}$}}       & 36.72                    & {\ul 12.78}              & 35.17                    & 0.70          & 0.55          & 0.43          & 0.33          & 58.3                        & -1.96                          & 170           \\
\multicolumn{1}{l|}{\textbf{$LongStory^{\neg C}$}}     & 35.44                    & 12.35                    & 33.89                    & 0.74          & 0.62          & 0.51          & 0.46          & 64.51                       & 0.23                           & 170           \\
\multicolumn{1}{l|}{\textbf{$LongStory^{\neg D}$}} & \textbf{37.03}           & \textbf{12.81}           & \textbf{35.35}           & {\ul 0.69}    & {\ul 0.54}    & {\ul 0.41}    & {\ul 0.32}    & 61.14                       & {\ul 6.38}                     & 170           \\ \hline
\multicolumn{11}{c}{\cellcolor[HTML]{EFEFEF}{\color[HTML]{3166FF} \textbf{30 Paragraphs}}}                                                                                                                                                                                      \\ \hline
\multicolumn{1}{l|}{}                           & \textbf{R-1}             & \textbf{R-2}             & \textbf{R-L}             & \textbf{B2}   & \textbf{B-3}  & \textbf{B-4}  & \textbf{B-5}  & \textit{\textbf{Coherence}} & \textit{\textbf{Completeness}} & \textbf{AvgL} \\ \hline
\multicolumn{1}{l|}{\textbf{Golden Label}}      & \cellcolor[HTML]{656565} & \cellcolor[HTML]{656565} & \cellcolor[HTML]{656565} & 0.48          & 0.25          & 0.11          & 0.04          & 80.19                       & 58.64                          & 164           \\ \hline
\multicolumn{1}{l|}{\textbf{Plotmachine}}       & 28.85                    & 8.89                     & 27.58                    & 0.75          & 0.62          & 0.49          & 0.38          & 63.24                       & 6.7                            & 193           \\
\multicolumn{1}{l|}{\textbf{BART}}              & 30.62                    & 10.53                    & 29.25                    & 0.78          & 0.67          & 0.55          & 0.46          & 61.65                       & -0.61                          & 164           \\
\multicolumn{1}{l|}{\textbf{GPT-2}}             & 29.43                    & 9.78                     & 28.06                    & 0.76          & 0.62          & 0.49          & 0.37          & \textbf{65.13}              & -0.26                          & 199           \\ \hline
\multicolumn{1}{l|}{\textbf{LongStory}}              & {\ul 32.24}              & \textbf{10.97}           & {\ul 30.90}              & \textbf{0.73} & \textbf{0.58} & \textbf{0.45} & \textbf{0.35} & {\ul 65}                    & \textbf{23.9}                  & 163           \\
\multicolumn{1}{l|}{\textbf{$LongStory^{\neg M}$}}       & 31.81                    & {\ul 10.96}              & 30.23                    & 0.74          & 0.60          & 0.47          & {\ul 0.37}    & 57.76                       & {\ul 19.43}                    & 167           \\
\multicolumn{1}{l|}{\textbf{$LongStory^{\neg C}$}}     & 29.96                    & 10.16                    & 28.76                    & 0.79          & 0.67          & 0.56          & 0.46          & 64.32                       & -0.1                           & 165           \\
\multicolumn{1}{l|}{\textbf{$LongStory^{\neg D}$}} & \textbf{32.42}           & 10.95                    & \textbf{31.02}           & {\ul 0.73}    & {\ul 0.59}    & {\ul 0.47}    & 0.48          & 60.4                        & 11.58                          & 168           \\ \hline
\multicolumn{11}{c}{\cellcolor[HTML]{EFEFEF}{\color[HTML]{3166FF} \textbf{50 Paragraph}}}                                                                                                                                                                                       \\ \hline
\multicolumn{1}{l|}{}                           & \textbf{R-1}             & \textbf{R-2}             & \textbf{R-L}             & \textbf{B2}   & \textbf{B-3}  & \textbf{B-4}  & \textbf{B-5}  & \textit{\textbf{Coherence}} & \textit{\textbf{Completeness}} & \textbf{AvgL} \\ \cline{1-3} \cline{5-11} 
\multicolumn{1}{l|}{\textbf{Golden Label}}      & \cellcolor[HTML]{656565} & \cellcolor[HTML]{656565} & \cellcolor[HTML]{656565} & 0.53          & 0.30          & 0.15          & 0.06          & 78.06                       & 52.4                           & 166           \\ \hline
\multicolumn{1}{l|}{\textbf{Plotmachine}}       & 29.01                    & 9.92                     & 28.14                    & {\ul 0.81}    & 0.70          & 0.59          & 0.49          & {\ul 67.95}                 & -9.08                          & 167           \\
\multicolumn{1}{l|}{\textbf{BART}}              & 27.83                    & 10.07                    & 27.05                    & 0.84          & 0.75          & 0.65          & 0.56          & 66.73                       & -5.66                          & 167           \\
\multicolumn{1}{l|}{\textbf{GPT-2}}             & 28.03                    & 9.94                     & 27.17                    & 0.84          & 0.73          & 0.62          & 0.52          & 67.73                       & 3.9                            & 198           \\ \hline
\multicolumn{1}{l|}{\textbf{LongStory}}              & \textbf{31.81}           & \textbf{11.65}           & \textbf{30.87}           & \textbf{0.79} & \textbf{0.66} & \textbf{0.54} & \textbf{0.43} & 63.67                       & \textbf{31.76}                 & 161           \\
\multicolumn{1}{l|}{\textbf{$LongStory^{\neg M}$}}       & 30.97                    & {\ul 11.51}              & 29.84                    & 0.81          & {\ul 0.69}    & 0.58          & {\ul 0.48}    & 63.06                       & -10.47                         & 168           \\
\multicolumn{1}{l|}{\textbf{$LongStory^{\neg C}$}}     & 27.87                    & 9.94                     & 27.12                    & 0.84          & 0.74          & 0.64          & 0.54          & \textbf{68.99}              & 20                             & 170           \\
\multicolumn{1}{l|}{\textbf{$LongStory^{\neg D}$}} & {\ul 30.99}              & 10.80                    & {\ul 30.09}              & 0.81          & 0.69          & {\ul 0.58}    & 0.48          & 63.25                       & 21.68                          & 167           \\ \hline
\end{tabular} % 
\vspace{5px}
}
\caption{ROUGE-1,2,L scores, in-self-BLEU n-gram scores, Coherence, Completeness scores, and the average length of a paragraph per 5,10,20,30,50 number of paragraphs. These are results from the combined test datasets of Writing Prompts, Reedsy Prompts, and Booksum. We bold the highest model scores in each column and underline the second highest model scores.}\label{tab2}
\end{table}
\vspace{-10px}
\subsection{Experimental Results}
\vspace{-5px}
\subsubsection{Coherence and Completeness.}
For the Coherence scorer, we divide two consecutive paragraphs into true pairs and two randomly selected paragraphs into fake pairs. We then train GPT-2 to use these constructed datasets on a classification task to distinguish whether the two paragraphs are true or fake folds\cite{Wang_2020,Guan_2020}, and convert the result of it into a score $\in [0, 1]$ using sigmoid function, as a Coherence score. We do the same for Completeness scorer, except for defining the true fold with the last paragraphs and the fake fold with the non-last paragraphs. It's important to highlight that the reported Completeness scores are calculated as the \textit{last paragraph's Completeness score} minus the \textit{average Completeness scores of non-last paragraphs}. This approach is chosen because an ideal story should generate a last paragraph appropriately and treat the body distinctly.
Regarding Coherence, as shown in table ~\ref{tab2}, our model is the best or second best in the 5,10,19,30 paragraph length sample. Plotmachine had a higher Coherence score than BART but lower than our model or GPT-2. In the ablated analysis, LongStory$^{\neg C}$ almost always outperforms LongStory$^{\neg M}$ and LongStory$^{\neg D}$ on Coherence. This suggests that $Memory$ is central to capturing the natural flow between paragraphs.
In the Completeness, our model is the best for all lengths. LongStory$^{\neg M}$ and LongStory$^{\neg C}$ have no absolute advantage in this column, so it is difficult to know exactly how $Memory$ and $Cheating$ affect Completeness. Still, since our model performs the best, we can say that a balanced mix of the two contexts improves performance. The results for LongStory$^{\neg D}$ also suggest that the newly added discourse tokens help improve Completeness.
\vspace{-15px}
\subsubsection{Relevance.}
We also show the average ROUGE scores between the generated documents and the golden labels to evaluate how well the model reproduces the golden label from the keywords. This serves as a metric to assess the relevance of the keywords to the predictions.
As shown in table ~\ref{tab2}, our model is best for results with 50 paragraph lengths and second best in many cases. Our model outperforms Plotmachine, GPT-2, and BART for all lengths of results. In the ablated analysis, LongStory$^{\neg D}$ scores the best, for the most part. This shows that the newly added order tokens have little or even a negative effect on relevance. However, in the repeatability test in the next section, LongStory$^{\neg D}$ performed worse than our model, which suggests that it reproduced the overall repetitive n-gram. For the most part, LongStory$^{\neg M}$ performed better than LongStory$^{\neg C}$, which means that $Cheating$ was more effective than $Memory$ in relevance.
\vspace{-15px}
\subsubsection{Repetitiveness.}
To see how much repetition occurs within a single output, we calculate the n-gram BLEU score by taking one paragraph as a hypothesis and the rest as a reference and averaging the results. We call the averaged results the in-SELF BLEU score\footnote{Note that in-self-BLEU score is not the same as self-BLEU score\cite{Rashkin_2020}. The self-BLEU score has taken one whole generated document as a hypothesis and the others as references, which cannot represent inner repetitiveness.}. The higher in-SELF BLEU score means that there are greater repetitions within the text and less diversity.
As shown in table ~\ref{tab2}, Our model performs best on the in-self-BLEU score for all but five paragraphs length. For five paragraph lengths, Plotmachine performs best, and ours is second highest. However, for longer results, our model is always better. In the ablated analysis, LongStory$^{\neg M}$ outperforms LongStory$^{\neg C}$ and BART. This suggests that $Cheating$ prevented the model from over-attending to past context or repeating n-grams irrelevant to the overall context. The better performance of our model over LongStory$^{\neg D}$ suggests that the order tokens we added positively affected reducing repetition.
\vspace{-10px}
\subsection{Further Analysis}
% \vspace{-10px}
\subsubsection{Zero-shot test.}
To see how well LongStory can have representations of the others, we do zero-shot tests on three models: WP,BK, and RP. As shown in table \ref{tab3}, Writing Prompt is large enough to have some representation on other types of test datasets, while Booksum and ReedsyPrompts are not. Total outperforms all others, underscoring the effectiveness of our methodology in integrating these diverse datasets.
\begin{table}[htb]
\centering
\resizebox{0.8\textwidth}{!}{%
\begin{tabular}{l|ccc|ccc|ccc}
\hline
                      & \multicolumn{3}{c|}{\textbf{Writing Prompts}}                                        & \multicolumn{3}{c|}{\textbf{Booksum}}                                                & \multicolumn{3}{c}{\textbf{Reedsy Prompts}}                                          \\ \hline
                      & \multicolumn{1}{c|}{\textbf{RP}} & \multicolumn{1}{c|}{\textbf{BK}} & \textbf{Total}  & \multicolumn{1}{c|}{\textbf{WP}} & \multicolumn{1}{c|}{\textbf{RP}} & \textbf{Total}  & \multicolumn{1}{c|}{\textbf{WP}} & \multicolumn{1}{c|}{\textbf{BK}} & \textbf{Total}  \\ \hline
Coherence    & 38.55                            & 43.75                            & \textbf{65.78} & 56.84                            & 48.62                            & \textbf{58.70} & 57.56                            & 47.79                            & \textbf{62.11} \\
Completeness & -0.1                             & 1.51                             & \textbf{27.06} & 1.67                             & -4.03                            & \textbf{12.37} & 7.75                             & 0.93                             & \textbf{21.29} \\ \hline
\end{tabular}
}
\vspace{5px}
\caption{Zero-shot test results averaged across test datasets of 5, 10, 20, 30, and 50 paragraphs. WP, RP, and BK are our models trained on Writing Prompts, Reedsy Prompts, and Booksum only, respectively. Total is the model trained on all three datasets combined. We bold the best results for each dataset.}\label{tab3}
\end{table}

\subsubsection{Augmented CWC.}\label{4.3.2}
We experiment with calibrating not only memory and cheating but also the proportion of input embeddings in the attention block. Table \ref{tab4} shows that the augmented version is better than ours on ROUGE-1,2, L scores but worse on Coherence and Completeness. This indicates that models excelling in relevance may not necessarily outperform in natural flow.

% \vspace{-45px}
\begin{table}[htb]
\centering
\begin{tabular}{l|lllcc}
\hline
         & \multicolumn{1}{c}{\textbf{R-1}} & \multicolumn{1}{c}{\textbf{R-2}} & \multicolumn{1}{c}{\textbf{R-L}} & \multicolumn{1}{l}{Coherence} & \multicolumn{1}{l}{Completeness} \\ \hline
\textbf{Aug-LongStory} & \textbf{38.25}                   & \textbf{13.06}                   & \textbf{36.65}                   & 61.01                                  & 5.25                                      \\
\textbf{LongStory }    & 38.04                            & 12.90                            & 36.46                            & \textbf{64.92}                         & \textbf{18.36}                            \\ \hline
\end{tabular}
\vspace{5px}
\caption{A comparison of the Augmented version of our model with the averaged scores of ROUGE-1, ROUGE-2, ROUGE-L, Coherence, and Completeness across test datasets of 5, 10, 20, 30, and 50 paragraphs. We bold the best results for each dataset.}\label{tab4}
\end{table}
% \vspace{-35px}
\vspace{-10px}
\section{Conclusion}
\vspace{-10px}
We introduce a novel task, Length Controllable Story Generation, aimed at recursively producing paragraph-by-paragraph stories of considerable length while ensuring coherence and completeness. To achieve this, we employ the long- and short-term contexts weight calibrator (CWC) and long story structural positions (LSP). The model is trained on three distinct datasets with varying average lengths, enabling it to learn representations of different lengths. Quantitative analysis demonstrates that our model excels in generating longer stories that exhibit coherence, completeness, relevance, and reduced repetitiveness compared to other baseline models.
% \vspace{-15px}
\section{Acknowledgements}
We thank anonymous reviewers for their constructive and insightful comments.
K. Jung is with ASRI, Seoul National University, Korea.
This work was supported by Institute of Information \& communications Technology Planning \&
Evaluation (IITP) grant funded by the Korea government(MSIT) [NO.2022-0-00184, Development and Study of AI Technologies to Inexpensively Conform to Evolving Policy on Ethics \& NO.2021-0-01343, Artificial Intelligence Graduate School Program (Seoul National University)]

\begin{thebibliography}{8}
\vspace{-10px}

\bibitem{Beltagy_2020}
Beltagy, I., Peters, M. E., \& Cohan, A. (2020). Longformer: The long-document transformer.arXiv preprint arXiv:2004.05150.
\bibitem{Lin_2021}
Lin, Z., \& Riedl, M. O. (2021, October). Plug-and-blend: a framework for plug-and-play controllable story generation with sketches. InProceedings of the AAAI Conference on Artificial Intelligence and Interactive Digital Entertainment(Vol. 17, No. 1, pp. 58-65).
\bibitem{Fan_2018}
Fan, A., Lewis, M., \& Dauphin, Y. (2018). Hierarchical neural story generation.arXiv preprint arXiv:1805.04833.
\bibitem{OpenAI_2023}
OpenAI. (2023). GPT-4 Technical Report.
\bibitem{Peng_2018}
Peng, N., Ghazvininejad, M., May, J., \& Knight, K. (2018, June). Towards controllable story generation. InProceedings of the First Workshop on Storytelling(pp. 43-49).

\bibitem{Rashkin_2020}
Rashkin, H., Celikyilmaz, A., Choi, Y., \& Gao, J. (2020). Plotmachines: Outline-conditioned generation with dynamic plot state tracking.arXiv preprint arXiv:2004.14967.
\bibitem{Yang_2022_Re3}
Yang, K., Peng, N., Tian, Y., \& Klein, D. (2022). Re3: Generating longer stories with recursive reprompting and revision.arXiv preprint arXiv:2210.06774.
\bibitem{Tang_2022_Etrica}
Tang, C., Lin, C., Huang, H., Guerin, F., \& Zhang, Z. (2022). EtriCA: Event-Triggered Context-Aware Story Generation Augmented by Cross Attention.arXiv preprint arXiv:2210.12463.
\bibitem{Brown_2020}
Brown, T., Mann, B., Ryder, N., Subbiah, M., Kaplan, J. D., Dhariwal, P., ... \& Amodei, D. (2020). Language models are few-shot learners.Advances in neural information processing systems,33, 1877-1901.
\bibitem{Lewis_2019}
Lewis, M., Liu, Y., Goyal, N., Ghazvininejad, M., Mohamed, A., Levy, O., ... \& Zettlemoyer, L. (2019). Bart: Denoising sequence-to-sequence pre-training for natural language generation, translation, and comprehension.arXiv preprint arXiv:1910.13461.
\bibitem{Rose_2010}
Rose, S., Engel, D., Cramer, N., \& Cowley, W. (2010). Automatic keyword extraction from individual documents. Text mining: applications and theory, 1-20.
\bibitem{Wang_2020}
Wang, S., Durrett, G., \& Erk, K. (2020). Narrative interpolation for generating and understanding stories.arXiv preprint arXiv:2008.07466.
\bibitem{Yang_2022_DOC}
Yang, K., Klein, D., Peng, N., \& Tian, Y. (2022). DOC: Improving Long Story Coherence With Detailed Outline Control.arXiv preprint arXiv:2212.10077.

\bibitem{Kryściński_2021}
Kryściński, W., Rajani, N., Agarwal, D., Xiong, C., \& Radev, D. (2021). Booksum: A collection of datasets for long-form narrative summarization.arXiv preprint arXiv:2105.08209.

\bibitem{Yao_2019}
Yao, L., Peng, N., Weischedel, R., Knight, K., Zhao, D., \& Yan, R. (2019, July). Plan-and-write: Towards better automatic storytelling. InProceedings of the AAAI Conference on Artificial Intelligence(Vol. 33, No. 01, pp. 7378-7385).
\bibitem{Alabdulkarim_2021_goal}
Alabdulkarim, A., Li, W., Martin, L. J., \& Riedl, M. O. (2021). Goal-directed story generation: Augmenting generative language models with reinforcement learning.arXiv preprint arXiv:2112.08593.
\bibitem{Pradyumna_2019}
Pradyumna, T., Murtaza, D., Lara, J. M., Mehta, A., \& Harrison, B. (2019). Controllable neural story plot generation via reward shaping. InProc. Int. Joint Conf. Artificial Intelligence(pp. 5982-5988).
\bibitem{Guan_2020}
Guan, J., Huang, F., Zhao, Z., Zhu, X., \& Huang, M. (2020). A knowledge-enhanced pretraining model for commonsense story generation.Transactions of the Association for Computational Linguistics,8, 93-108.
\bibitem{Peng_2021}
Peng, X., Li, S., Wiegreffe, S., \& Riedl, M. (2021). Inferring the reader: Guiding automated story generation with commonsense reasoning.arXiv preprint arXiv:2105.01311.


\bibitem{Vaswani_2017}
Vaswani, A., Shazeer, N., Parmar, N., Uszkoreit, J., Jones, L., Gomez, A. N., ... \& Polosukhin, I. (2017). Attention is all you need. Advances in neural information processing systems, 30.
\bibitem{Lin_2004}
Lin, C. Y. (2004, July). Rouge: A package for automatic evaluation of summaries. InText summarization branches out(pp. 74-81).
\bibitem{Papineni_2002}
Papineni, K., Roukos, S., Ward, T., \& Zhu, W. J. (2002, July). Bleu: a method for automatic evaluation of machine translation. InProceedings of the 40th annual meeting of the Association for Computational Linguistics(pp. 311-318).

\bibitem{Safovich_2020}
Safovich, Y., \& Azaria, A. (2020, November). Fiction sentence expansion and enhancement via focused objective and novelty curve sampling. In2020 IEEE 32nd International Conference on Tools with Artificial Intelligence (ICTAI)(pp. 835-843). IEEE.
\bibitem{Li_2019}
Li, J., Bing, L., Qiu, L., Chen, D., Zhao, D., \& Yan, R. (2019, July). Learning to write stories with thematic consistency and wording novelty. InProceedings of the AAAI Conference on Artificial Intelligence(Vol. 33, No. 01, pp. 1715-1722).
\bibitem{Hu_2022}
Hu, Z., Chan, H. P., Liu, J., Xiao, X., Wu, H., \& Huang, L. (2022). Planet: Dynamic content planning in autoregressive transformers for long-form text generation.arXiv preprint arXiv:2203.09100.
\bibitem{Yang_2021}
Yang, K., \& Klein, D. (2021). FUDGE: Controlled text generation with future discriminators.arXiv preprint arXiv:2104.05218.
\bibitem{Sakaguchi_2021}
Sakaguchi, K., Bhagavatula, C., Bras, R. L., Tandon, N., Clark, P., \& Choi, Y. (2021). proscript: Partially ordered scripts generation via pre-trained language models.arXiv preprint arXiv:2104.08251.
\bibitem{Budzianowski_2019}
Budzianowski, P., \& Vulić, I. (2019). Hello, it's GPT-2--how can I help you? towards the use of pretrained language models for task-oriented dialogue systems.arXiv preprint arXiv:1907.05774.


\bibitem{Welleck_2020}
Welleck, S., Kulikov, I., Kim, J., Pang, R. Y., \& Cho, K. (2020). Consistency of a recurrent language model with respect to incomplete decoding.arXiv preprint arXiv:2002.02492.
\bibitem{Zellers_2019}
Zellers, R., Holtzman, A., Rashkin, H., Bisk, Y., Farhadi, A., Roesner, F., \& Choi, Y. (2019). Defending against neural fake news. Advances in neural information processing systems, 32.
\bibitem{Guan_2021}
Guan, J., Mao, X., Fan, C., Liu, Z., Ding, W., \& Huang, M. (2021). Long text generation by modeling sentence-level and discourse-level coherence. arXiv preprint arXiv:2105.08963.
\bibitem{Lewis_2020}
Lewis, P., Perez, E., Piktus, A., Petroni, F., Karpukhin, V., Goyal, N., ... \& Kiela, D. (2020). Retrieval-augmented generation for knowledge-intensive nlp tasks. Advances in Neural Information Processing Systems, 33, 9459-9474.
\bibitem{McCoy_2023}
McCoy, R. T., Smolensky, P., Linzen, T., Gao, J., \& Celikyilmaz, A. (2023). How much do language models copy from their training data? evaluating linguistic novelty in text generation using raven. Transactions of the Association for Computational Linguistics, 11, 652-670.
\end{thebibliography}
%

\end{document}